\documentclass{article}
\usepackage{ijcai21}

\usepackage{times}
\usepackage{soul}
\usepackage{url}
\usepackage[hidelinks]{hyperref}
\usepackage[utf8]{inputenc}
\usepackage[small]{caption}
\usepackage{graphicx}
\usepackage{amsmath}
\usepackage{booktabs}
\usepackage{hyperref}
\usepackage{url}

\usepackage{wrapfig,lipsum}
\usepackage{graphicx}
\usepackage{tabularx}
\usepackage{multirow}
\usepackage{amsmath}
\usepackage{amsthm}
\usepackage{amsfonts} 
\usepackage{amssymb}
\usepackage{mathrsfs}
\usepackage{fixmath}
\usepackage{mathtools}
\usepackage{threeparttable}
\usepackage{graphicx}
\usepackage{color}
\usepackage[linesnumbered]{algorithm2e}

\usepackage{hyperref}

\usepackage{caption}
\usepackage{subcaption}
\usepackage{enumitem}
\usepackage{bm}
\usepackage{comment}
\usepackage{makecell}

\usepackage{etoolbox}
\patchcmd{\thebibliography}
  {\settowidth}
  {\setlength{\parsep}{-0.5pt}\setlength{\itemsep}{0pt plus 0pt}\settowidth}
  {}{}
\apptocmd{\thebibliography}
  {\small}
  {}{}

\usepackage[switch]{lineno}

\newtheorem{definition}{Definition}
\newtheorem{prop}{Proposition}
\newtheorem{lemma}{Lemma}

\newcommand{\xhdr}[1]{{\noindent\bfseries #1}.}

\newcommand{\jure}[1]{{{\textcolor{red} {[{Jure:}  #1]}}}}

\newcommand{\namelong}{Multi-hop Attention Graph Neural Network\xspace}
\newcommand{\name}{MAGNA\xspace}

\newcommand{\cut}[1]{}
\newcommand{\hide}[1]{}

\title{Multi-hop Attention Graph Neural Networks}

\author{
Guangtao Wang$^1$\footnote{Contact Author, xjtuwgt@gmail.com}\and
Rex Ying$^2$\and
Jing Huang$^{1}$\And
Jure Leskovec$^2$\\
\affiliations
$^1$JD AI Research\\
$^2$Computer Science, Stanford University\\
\emails
guangtao.wang@jd.com, rexying@stanford.edu,
jing.huang@jd.com,
jure@cs.stanford.edu
}

\begin{document}
\maketitle

\begin{abstract}
Self-attention mechanism in graph neural networks (GNNs) led to state-of-the-art performance on many graph representation learning tasks. Currently, at every layer, a node computes attention independently for each of its graph neighbors. 
However, such attention mechanism is limited as it does not consider nodes that are not connected by an edge but can provide important network context information.
%
Here we propose {\em Multi-hop Attention Graph Neural Network (\name)}, a principled way to incorporate multi-hop context information into every layer of GNN attention computation.
\name diffuses the attention scores across the network, which increases the ``receptive field'' for every layer of the GNN.
Unlike previous approaches, \name uses a diffusion prior on attention values, to efficiently account for all paths between the pair of not-connected nodes. We demonstrate theoretically and experimentally that \name captures large-scale structural information in every layer, and has a low-pass effect that eliminates noisy high-frequency information from the graph. 
Experimental results on node classification as well as knowledge graph completion benchmarks show that \name achieves state-of-the-art results: \name achieves up to $5.7\%$ relative error reduction over the previous state-of-the-art on Cora, Citeseer, and Pubmed. \name also obtains strong performance on a large-scale Open Graph Benchmark dataset. Finally, on knowledge graph completion \name advances state-of-the-art on WN18RR and FB15k-237 across four different performance metrics.
\end{abstract}

\section{Introduction}

Self-attention~\cite{Bahdanau2015ICLR,vaswani2017attention} has pushed the state-of-the-art in many domains including graph presentation learning~\cite{devlin2019bert}. 
Graph Attention Network (GAT)~\cite{velivckovic2017graph} and related models~\cite{li2018deeper,wang2019improving,liu2019geniepath,oono2020graph} developed attention mechanism for Graph Neural Networks (GNNs), which compute attention scores between nodes connected by an edge, allowing the model to attend to messages of node's neighbors.


\begin{figure}[!t]
    \centering
    \includegraphics[width=0.35\textwidth]{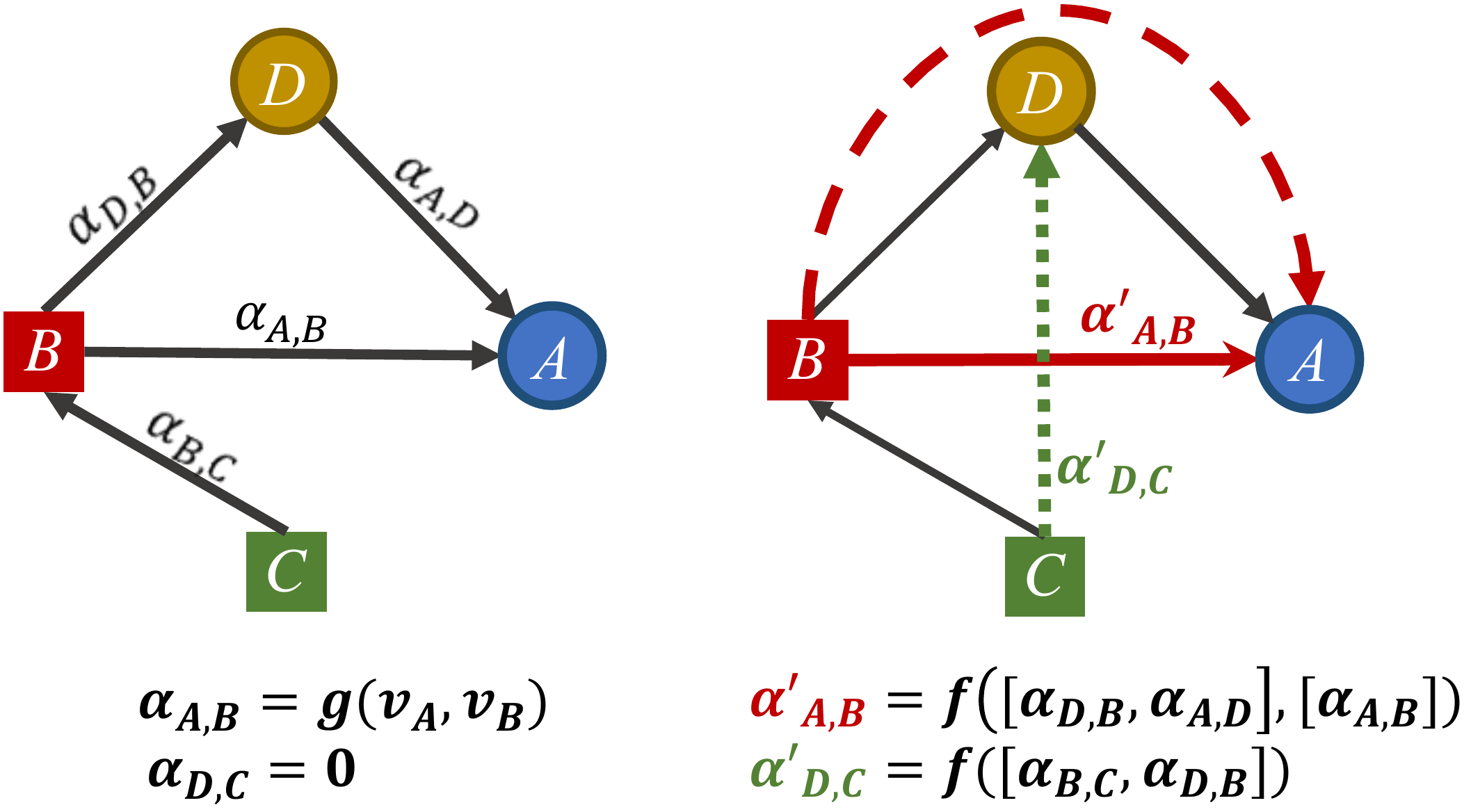}
    \caption{{\bf Multi-hop attention diffusion.} Consider making a prediction at nodes $A$ and $D$.
    \textbf{Left}: A single GAT layer computes attention scores $\mathbold{\alpha}$ between directly connected pairs of nodes (i.e., edges) and thus $\mathbold{\alpha}_{D,C}=0$. Furthermore, the attention $\mathbold{\alpha}_{A,B}$ between $A$ and $B$ only depends on node representations of $A$ and $B$. \textbf{Right}: A single \name layer: (1) captures the information of $D$'s two-hop neighbor node $C$ via multi-hop attention $\mathbold{\alpha'}_{D,C}$; and (2) enhances graph structure learning by considering all paths between nodes via diffused attention, which is 
    based on powers of graph adjacency matrix. \name makes use of node $D$'s features for attention computation between $A$ and $B$. This means that two-hop attention in \name is context (node $D$) dependent.
    }\label{fig:intro}
\end{figure}

However, such attention computation on pairs of nodes connected by edges implies that a node can only attend to its immediate neighbors to compute its (next layer) representation. This implies that receptive field of a single GNN layer is restricted to one-hop network neighborhoods. Although stacking multiple GAT layers could in principle enlarge the receptive field and learn non-neighboring interactions, such deep GAT architectures suffer from the oversmoothing problem~\cite{wang2019improving,liu2019geniepath,oono2020graph} and do not perform well. Furthermore, edge attention in a single GAT layer is based solely on representations of the two nodes at the edge endpoints, and does not depend on their graph neighborhood context. In other words, the one-hop attention mechanism in GATs limits their ability to explore the relationship between the broader graph structure.
While previous works~\cite{xu2018representation,klicpera2019diffusion} have shown advantages in performing multi-hop message-passing in a single layer, these approaches are not graph-attention based. Therefore, incorporating multi-hop neighboring context into the attention computation in graph neural networks remains to be explored. 


Here we present {\em \namelong (\name)}, an effective multi-hop self-attention mechanism 
for graph structured data. \name uses a novel graph attention diffusion layer (Figure \ref{fig:intro}), where we first compute attention weights on edges (represented by solid arrows), and then compute self-attention weights (dotted arrows) between disconnected pairs of nodes through an attention diffusion process using the attention weights on the edges.

Our model has two main advantages:
(1) \name captures long-range interactions between nodes that are not directly connected but may be multiple hops away. Thus the model enables effective long-range message passing, from important nodes multiple hops away.
(2) The attention computation in \name is context-dependent. The attention value in GATs~\cite{velivckovic2017graph} only depends on node representations of the previous layer, and is zero between non-connected pairs of nodes. In contrast, for any pair of nodes within a chosen multi-hop neighborhood, \name computes attention by aggregating the attention scores over all the possible paths (length $\geq 1$) connecting the two nodes. 


Mathematically we show that \name places a Personalized Page Rank (PPR) prior on the attention values. We further apply spectral graph analysis to show that \name emphasizes on large-scale graph structure and lowering high-frequency noise in graphs. Specifically, \name enlarges the lower Laplacian eigen-values, which correspond to the large-scale structure in the graph, and suppresses the higher Laplacian eigen-values which correspond to more noisy and fine-grained information in the graph.


We experiment on standard datasets for semi-supervised node classification as well as knowledge graph completion. 
Experiments show that \name achieves state-of-the-art results: \name achieves up to $5.7\%$ relative error reduction over previous state-of-the-art on Cora, Citeseer, and Pubmed. \name also obtains better performance on a large-scale Open Graph Benchmark dataset. On knowledge graph completion, \name advances state-of-the-art on WN18RR and FB15k-237 across four metrics, with the largest gain of 7.1\% in the metric of Hit at 1.


Furthermore, we show that \name with just 3 layers and 6 hop wide attention per layer significantly out-performs GAT with 18 layers, even though both architectures have the same receptive field.
Moreover, our ablation study reveals the synergistic effect of the essential components of \name, including layer normalization and multi-hop diffused attention.
We further observe that compared to GAT, the attention values learned by \name have higher diversity, indicating the ability to better pay attention to important nodes.

\section{Multi-hop Attention Graph Neural Network (\name)}
We first discuss the background and explain the novel multi-hop attention diffusion module and the \name architecture. 


\subsection{Preliminaries}
Let $\mathcal{G} = (\mathcal{V}, \mathcal{E})$ be a given graph, where $\mathcal{V}$ is the set of $N_n$ nodes, $\mathcal{E}  \subseteq \mathcal{V} \times \mathcal{V}$ is the set of $N_{e}$ edges connecting $M$ pairs of nodes in $\mathcal{V}$. Each node $v \in \mathcal{V}$ and each edge $e \in \mathcal{E}$ are associated with their type mapping functions: $\phi:$ $\mathcal{V} \rightarrow \mathcal{T}$ and $\psi:$  $\mathcal{E} \rightarrow \mathcal{R}$. Here $\mathcal{T}$ and $\mathcal{R}$ denote the sets of node types (labels) and edge/relation types.
Our framework supports learning on heterogeneous graphs with multiple elements in $\mathcal{R}$.

A general Graph Neural Network (GNN) approach learns an embedding that maps nodes and/or edge types into a continuous vector space. Let $\mathbold{X} \in \mathbb{R}^{N_n\times d_n}$ and $\mathbold{R} \in \mathbb{R}^{N_r\times d_r}$ be the node embedding and edge/relation type embedding, where $N_n = |\mathcal{V}|$, $N_r = |\mathcal{R}|$, $d_n$ and $d_r$ represent the embedding dimension of node and edge/relation types, each row $\bm{x}_{i}=\mathbold{X}[i:]$ represents the embedding of node $v_i$ ($1\leq i \leq N_n$), and $\bm{r}_{j}=\mathbold{R}[j:]$ represents the embedding of relation $r_j$ ($1\leq j \leq N_r$). 
\name builds on GNNs, while bringing together the benefits of Graph Attention and Diffusion techniques.




\subsection{Multi-hop Attention Diffusion}
\label{sec:att_diffusion}

We first introduce attention diffusion to compute the multi-hop attention directly, which operates on the \name's attention scores at each layer. The input to the attention diffusion operator is a set of triples $(v_i, r_k, v_j)$, where $v_i, v_j$ are nodes and $r_k$ is the edge type. \name first computes the attention scores on all edges. The attention diffusion module then computes the attention values between pairs of nodes that are not directly connected by an edge, based on the edge attention scores, via a diffusion process.
The attention diffusion module can then be used as a component in \name architecture, which we will further elaborate in Section~\ref{sec:arch}. 

\paragraph{Edge Attention Computation} 
At each layer $l$, a vector message is computed for each triple $(v_i, r_k, v_j)$. To compute the representation of $v_j$ at layer $l+1$, all messages from triples incident to $v_j$ are aggregated into a single message, which is then used to update $v_j^{l+1}$. 

In the first stage, the attention score $s$ of an edge $(v_i, r_k, v_j)$ is computed by the following:
\begin{equation}\label{eq:attention}
    s^{(l)}_{i, k, j} = \delta(\mathbold{v}^{(l)}_{a}\tanh(\mathbold{W}^{(l)}_{h}\mathbold{h^{(l)}_{i}}\Vert\mathbold{W}^{(l)}_{t}\mathbold{h^{(l)}_{j}}\Vert \mathbold{W}^{(l)}_r\mathbold{r}_{k}))
\end{equation}
where $\delta=\text{LeakyReLU}$, $\mathbold{W}^{(l)}_h$, $\mathbold{W}^{(l)}_t$$\in$ $\mathbb{R}^{ d^{(l)}\times d^{(l)}}$, 
$\mathbold{W}^{(l)}_r$$\in$ $\mathbb{R}^{d^{(l)}\times d_r}$
and $\mathbold{v}^{(l)}_a$ $\in$ $\mathbb{R}^{1\times 3d^{(l)}}$ are the trainable weights shared by $l$-th layer. $\mathbold{h}_{i}^{(l)} \in \mathbb{R}^{d^{(l)}}$ represents the embedding of node $i$ at $l$-th layer, and $\mathbold{h}_{i}^{(0)}=\mathbold{x}_{i}$. $\mathbold{r}_{k}$ is the trainable relation embedding of the $k$-th relation type $r_k$ ($1\leq k \leq N_r$), and $a\Vert b$ denotes concatenation of embedding vectors $a$ and $b$.
For graphs with no relation type, we treat as a degenerate categorical distribution with 1 category\footnote{In this case, we can view that there is only one ``pseudo'' relation type (category), i.e., $N_r$ = 1}.

Applying Eq. \ref{eq:attention} on each edge of the graph $\mathcal{G}$, we obtain an attention score matrix $\mathbold{S}^{(l)}$:
\begin{equation}
    \mathbold{S}^{(l)}_{i,j} = 
    \begin{cases}
        s^{(l)}_{i, k, j}, & \text{if } (v_i, r_k, v_j) \text{ appears in } \mathcal{G}\\
        -\infty,              & \text{otherwise}
    \end{cases}
\end{equation}
Subsequently we obtain the attention matrix $\mathbold{A}^{(l)}$ by performing row-wised softmax over the score matrix $\mathbold{S}^{(l)}$:
   $\mathbold{A}^{(l)} = \text{softmax}(\mathbold{S^{(l)}})$.
 $\mathbold{A}^{(l)}_{ij}$ denotes the attention value at layer $l$ when aggregating message from node $j$ to node $i$.

\paragraph{Attention Diffusion for Multi-hop Neighbors} 
In the second stage, we further enable attention between nodes that are not directly connected in the network. We achieve this via the following attention diffusion procedure.
The procedure computes the attention scores of multi-hop neighbors via graph diffusion based on the powers of the 1-hop attention matrix $\mathbold{A}$:
%
\begin{equation}\label{eq:attentionDiffusion}
    \begin{aligned}
            \mathcal{A} & = \sum_{i=0}^{\infty}\theta_{i}\mathbold{A}^{i} \text{ where}  
            \sum_{i=0}^{\infty}\theta_{i} = 1 \text{ and } \theta_{i} > 0
    \end{aligned}
\end{equation}
where $\theta_i$ is the attention decay factor and $\theta_i > \theta_{i+1}$. 
The powers of attention matrix, ${A}^{i}$, give us the number of relation paths from node $h$ to node $t$ of length up to $i$, increasing the receptive field of the attention (Figure~\ref{fig:intro}). Importantly, the mechanism allows the attention between two nodes to not only depend on their previous layer representations, but also taking into account of the paths between the nodes, effectively creating attention shortcuts between nodes that are not directly connected (Figure \ref{fig:intro}). Attention through each path is also weighted differently, depending on $\theta$ and the path length.

In our implementation we utilize the geometric distribution $\theta_i$ = $\alpha(1-\alpha)^{i}$, where $\alpha \in (0, 1]$. The choice is based on the inductive bias that nodes further away should be weighted less in message aggregation, and nodes with different relation path lengths to the target node are sequentially weighted in an independent manner. 
In addition, notice that if we define $\theta_{0}$ = $\alpha \in (0, 1]$, $\mathbold{A}^{0}= \mathbold{I}$, then Eq. \ref{eq:attentionDiffusion} gives the Personalized Page Rank~(PPR) procedure on the graph with the attention matrix $\mathbold{A}$ and teleport probability $\alpha$. Hence the diffused attention weights, $\mathcal{A}_{ij}$, can be seen as the influence of node $j$ to node $i$. We further elaborate the significance of this observation in Section \ref{sec:analysis}.

We can also view $\mathcal{A}_{ij}$ as the attention value of node $j$ to $i$ since $\sum_{j=1}^{N_{n}} \mathcal{A}_{ij} = 1$.\footnote{Obtained by the definition $\mathbold{A}^{(l)} = \text{softmax}(\mathbold{S^{(l)}})$ and Eq. \ref{eq:attentionDiffusion}.}
We then define the \emph{graph attention diffusion} based feature aggregation as
\begin{equation}
\label{eq:diffusion}
    \text{AttDiff}(\mathcal{G},\mathbold{H}^{(l)}, \Theta) = \mathcal{A}\mathbold{H}^{(l)},
\end{equation}
where $\Theta$ is the set of parameters for computing attention. Thanks to the diffusion process defined in Eq. \ref{eq:attentionDiffusion}, \name uses the same number of parameters as if we were only computing attention between nodes connected via edges. This ensures runtime efficiency (refer to Appendix\footnote{Appendix can be found via \url{https://arxiv.org/abs/2009.14332}} A for complexity analysis) and model generalization.


\paragraph{Approximate Computation for Attention Diffusion} 
For large graphs computing the exact attention diffusion matrix $\mathcal{A}$ using Eq. \ref{eq:attentionDiffusion} may be prohibitively expensive, due to computing the powers of the attention matrix~\cite{klicpera2018predict}. To resolve this bottleneck, we proceed as follows:
Let $\mathbold{H}^{(l)}$ be the input entity embedding of the $l$-th layer ($\mathbold{H}^{(0)} = \mathbold{X}$) and $\theta_{i} = \alpha(1-\alpha)^{i}$.
Since \name only requires aggregation via $\mathcal{A}\mathbold{H}^{(l)}$,
we can approximate $\mathcal{A}\mathbold{H}^{(l)}$ by defining a sequence $Z^{(K)}$ which converges to the true value of $\mathcal{A}\mathbold{H}^{(l)}$ (Proposition \ref{prop:approxGAT}) as $K \rightarrow \infty$:
\begin{equation}\label{eq:approxGAT}
    \mathbold{Z}^{(0)} = \mathbold{H}^{(l)},
    \mathbold{Z}^{(k+1)} = (1 - \alpha)\mathbold{A}\mathbold{Z}^{(k)}+ \alpha\mathbold{Z}^{(0)},
\end{equation}
where $0 \leq k < K$.

\begin{prop}\label{prop:approxGAT}
    $\lim_{K\rightarrow\infty}Z^{(K)} = \mathcal{A}\mathbold{H}^{(l)}$ 
\end{prop}
In the Appendix we give the proof which relies on the expansion of Eq.~\ref{eq:approxGAT}.

Using the above approximation, the complexity of attention computation with diffusion is still $O(|E|)$, with a constant factor corresponding to the number of hops $K$. 
In practice, we find that choosing the values of 
$K$ such that $3\le K \le 10$ results in good model performance.
Many real-world graphs exhibit small-world property, in which case even a smaller value of $K$ is sufficient. For graphs with larger diameter, we choose larger $K$, and lower the value of $\alpha$.

\begin{figure}[!t]
    \centering
    \includegraphics[width=0.4\textwidth]{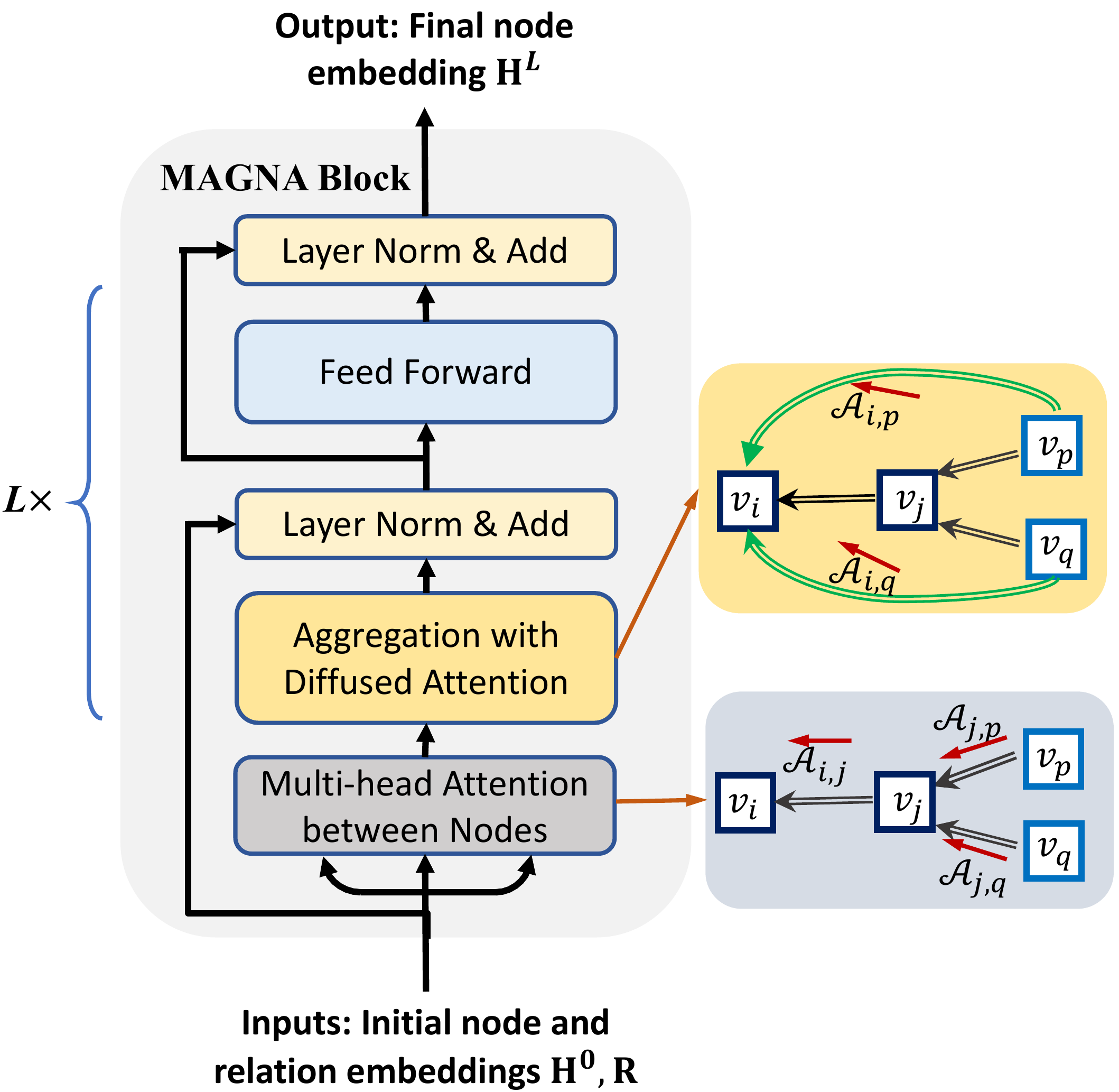} 
    \caption{{\bf \name Architecture}. Each \name block consists of attention computation, attention diffusion, layer normalization, feed forward layers, and 2 residual connections for each block. \name blocks can be stacked to constitute a deep model. As illustrated on the right, context-dependent attention is achieved via the attention diffusion process. Here $v_i, v_j, v_p, v_q\in\mathcal{V}$ are nodes in the graph.}
    \label{fig:arch}
\end{figure}

\subsection{Multi-hop Attention based GNN Architecture}\label{sec:arch}

Figure \ref{fig:arch} provides an architecture overview of the \name Block that can be stacked multiple times. 


\paragraph{Multi-head Graph Attention Diffusion Layer} Multi-head attention~\cite{vaswani2017attention,velivckovic2017graph} is used to 
allow the model to jointly attend to information from different representation
sub-spaces at different viewpoints. 
In Eq. \ref{eq:multihead_att}, the attention diffusion for each head $i$ is computed separately with Eq. \ref{eq:diffusion}, and aggregated:
\begin{equation}
\label{eq:multihead_att}
\begin{aligned}
    \hat{\mathbold{H}}^{(l)} & = \text{MultiHead}(\mathcal{G}, \mathbold{\tilde{H}}^{(l)}) = \left(\mathbin\Big\Vert_{i=1}^{M}\textrm{head}_i\right)\mathbold{W}_o \\
    \textrm{head}_i &= \text{AttDiff}(\mathcal{G}, \mathbold{\tilde{H}}^{(l)}, \Theta_{i}), \mathbold{\tilde{H}}^{(l)} = \text{LN}(\mathbold{H}^{(l)}),
\end{aligned}
\end{equation}
where $\mathbin\Vert$ denotes concatenation and $\Theta_{i}$ are the parameters in Eq.~\ref{eq:attention} for the $i$-th head ($1\leq i \leq M$), $\mathbold{W}_o$ represents a parameter matrix, and $\text{LN}$ = $\text{LayerNorm}$. Since we calculate the attention diffusion in a recursive way in Eq.~\ref{eq:approxGAT}, we add layer normalization which helpful to stabilize the recurrent computation procedure~\cite{ba2016layer}.

\paragraph{Deep Aggregation} Moreover our \name block contains a fully connected feed-forward sub-layer, which consists of a two-layer feed-forward network. We also add the layer normalization and residual connection in both sub-layers, allowing for a more expressive aggregation step for each block~\cite{xiong2020layer}:
\begin{equation}\label{eq:diffusionGAT}
\begin{aligned}
    \hat{\mathbold{H}}^{(l+1)} & =  \hat{\mathbold{H}}^{(l)} + \mathbold{H}^{(l)}\\ 
    \mathbold{H}^{(l+1)} & =  \mathbold{W}^{(l)}_{2}\text{ReLU}\left(\mathbold{W}^{(l)}_1\text{LN}(\hat{\mathbold{H}}^{(l+1)})\right) + \hat{\mathbold{H}}^{(l+1)}
\end{aligned}
\end{equation}
\paragraph{\name generalizes GAT} \name extends GAT via the diffusion process. The feature aggregation in GAT is $\mathbold{H}^{(l+1)} = \sigma(\mathbold{A}\mathbold{H}^{(l)}\mathbold{W}^{(l)})$, where $\sigma$ represents the activation function. We can divide GAT layer into two components as follows:
\begin{equation}\label{eq:gat}
    \mathbold{H}^{(l+1)} = 
    \underbrace{\sigma}_\textrm{(2)}(\underbrace{\mathbold{A}\mathbold{H}^{(l)}\mathbold{W}^{(l)}}_\textrm{(1)}).
\end{equation}
In component (1), 
\name removes the restriction of attending to direct neighbors, without requiring additional parameters as $\mathcal{A}$ is induced from $\mathbold{A}$. 
For component (2) \name uses layer normalization and deep aggregation to achieve higher expressive power compared to elu nonlinearity in GAT.

\section{Analysis of Graph Attention Diffusion}
\label{sec:analysis}
In this section, we investigate the benefits of \name from the viewpoint of discrete signal processing on graphs~\cite{sandryhaila2013discrete}. Our first result demonstrates that \name can better capture large-scale structural information. 
Our second result explores the relation between \name and Personalized PageRank (PPR). 



\subsection{Spectral Properties of Graph Attention Diffusion}
We view the attention matrix $\mathbold{A}$ of GAT, and $\mathcal{A}$ of \name as weighted adjacency matrices, and apply Graph Fourier transform and spectral analysis (details in  Appendix) to show the effect of \name as a graph low-pass filter, being able to more effectively capture large-scale structure in graphs. 
By Eq. \ref{eq:attentionDiffusion}, the sum of each row of either $\mathcal{A}$ or $\mathbold{A}$ is 1. Hence the normalized graph Laplacians are $\hat{\mathbold{L}}_{sym} = \mathbold{I} - \mathcal{A}$ and $\mathbold{L}_{sym} = \mathbold{I} - \mathbold{A}$ for $\mathcal{A}$ and $\mathbold{A}$ respectively. We can get the following proposition:
\begin{prop}
\label{prop:lowerpass} 
Let $\hat{\lambda}_{i}^{g}$ and $\lambda_{i}^{g}$ be the $i$-th eigeinvalues of $\hat{\mathbold{L}}_{sym}$ and $\mathbold{L}_{sym}$. 
 \begin{equation}\label{eq:lowpassFilter}
    \frac{\hat{\lambda}_{i}^{g}}{\lambda_{i}^{g}} = \frac{1 - \frac{\alpha}{1 - (1-\alpha)(1 - \lambda_{i}^{g})}}{\lambda_{i}^{g}} = \frac{1}{\frac{\alpha}{1 -\alpha} + \lambda_{i}^{g}}.
\end{equation}
\end{prop}
Refer to Appendix for the proof.
We additionally have $\lambda_{i}^{g} \in [0,2]$ (proved by~\cite{ng2002spectral}).
Eq. \ref{eq:lowpassFilter} shows that when $\lambda_{i}^{g}$ is small such that $\frac{\alpha}{1 -\alpha} + \lambda_{i}^{g} < 1$, then $\hat{\lambda}_{i}^{g} > \lambda_{i}^{g}$, whereas for large $\lambda_{i}^{g}$, $\hat{\lambda}_{i}^{g} < \lambda_{i}^{g}$. This relation indicates that the use of $\mathcal{A}$ increases smaller eigenvalues and decreases larger eigenvalues\footnote{The eigenvalues of $\mathcal{A}$ and $\mathbold{A}$ correspond to the same eigenvectors, as shown in Proposition 2 in Appendix.}. 
See Section \ref{sec:analysis} for its empirical evidence.
The low-pass effect increases with smaller $\alpha$.

The eigenvalues of the low-frequency signals describe the large-scale structure in the graph~\cite{ng2002spectral} and have been shown to be crucial in graph tasks~\cite{klicpera2019diffusion}. As $\lambda_{i}^{g} \in [0,2]$~\cite{ng2002spectral} and $\frac{\alpha}{1 -\alpha} > 0$, the reciprocal format in Eq.~\ref{eq:lowpassFilter} will amplify the ratio of lower eigenvalues to the sum of all eigenvalues. In contrast, high eigenvalues corresponding to noise are suppressed. 


\subsection{Personalized PageRank Meets Graph Attention Diffusion}\label{subsec:relatedRWR}

We can also view the attention matrix $\mathbold{A}$ as a random walk matrix on graph $\mathcal{G}$ since $\sum_{j=1}^{N_n} \mathbold{A}_{i,j}= 1$ and $\mathbold{A}_{i,j}>0$. If we perform Personalized PageRank (PPR) 
with parameter $\alpha \in (0,1]$ on $\mathcal{G}$ with transition matrix $\mathbold{A}$, the fully Personalized PageRank~\cite{lofgren2015efficient} is defined as:
\begin{equation}
    \mathbold{A}_{ppr} = \alpha(\mathbold{I} - (1 - \alpha) \mathbold{A})^{-1}
\end{equation}
Using the power series expansion for the matrix inverse, we obtain
\begin{equation}
    \mathbold{A}_{ppr} = \alpha\sum_{i=0}^{\infty}(1-\alpha)^{i} \mathbold{A}^{i} = \sum_{i=0}^{\infty} \alpha(1-\alpha)^{i} \mathbold{A}^{i}
\end{equation}

Comparing to the diffusion Equation \ref{eq:attentionDiffusion} with $\theta_{i} = \alpha(1-\alpha)^{i}$, we have the following proposition.

\begin{prop}
 Graph attention diffusion defines a personalized page rank with parameter $\alpha \in (0,1]$ on $\mathcal{G}$ with transition matrix $\mathbold{A}$, i.e., $\mathcal{A} = \mathbold{A}_{ppr}$.
\end{prop}

The parameter $\alpha$ in \name is equivalent to the teleport probability of PPR.
PPR provides a good relevance score between nodes in a weighted graph (the weights from the attention matrix $\mathbold{A}$). 
In summary, \name places a PPR prior over node pairwise attention scores:
the diffused attention between node $i$ and $j$ depends on the attention scores on the edges of all paths between $i$ and $j$.


\section{Experiments}
We evaluate \name on two classical tasks\footnote{All datasets are public. And our implementation is available at \url{https://github.com/xjtuwgt/GNN-MAGNA}}: (1) on node classification we achieve an average of $5.7\%$ relative error reduction; (2) on knowledge graph completion we achieve $7.1\%$ relative improvement in the Hit@1 metric.\footnote{Please see the definitions of these two tasks in Appendix.} 


\subsection{Task 1: Node Classification}

\paragraph{Datasets} We employ four benchmark datasets for node classification: (1) standard citation network benchmarks Cora, Citeseer and Pubmed~\cite{sen2008collective,kipf2016semi};
and (2) a benchmark dataset ogbn-arxiv on 170k nodes and 1.2m edges from the Open Graph Benchmark~\cite{hu2020ogb}.
We follow the standard data splits for all datasets. Further information about these datasets is summarized in the Appendix.

\paragraph{Baselines} We compare against a comprehensive suite of state-of-the-art GNN methods including: GCNs~\cite{kipf2016semi}, Chebyshev filter based GCNs~\cite{defferrard2016convolutional}, DualGCN~\cite{zhuang2018dual}, 
JKNet~\cite{xu2018representation},
LGCN~\cite{gao2018large}, 
Diffusion-GCN (Diff-GCN)~\cite{klicpera2019diffusion}, APPNP~\cite{klicpera2018predict}, Graph U-Nets (g-U-Nets)~\cite{gao2019graph}, and GAT~\cite{velivckovic2017graph}. 

\begin{table}[!t]
	\centering
	\resizebox{0.995\columnwidth}{!}{%
	\begin{threeparttable}
		\begin{tabularx}{0.66\textwidth}{c | l c c c} 
			\hline
			& Models & Cora & Citeseer & Pubmed\\
			\hline
			\multirow{8}{*}{\rotatebox{90}{\hspace*{+2pt} Baselines}} & 
            GCN~\cite{kipf2016semi} & 81.5  & 70.3 & 79.0 \\
			&Cheby~\cite{defferrard2016convolutional} & 81.2 &  69.8 & 74.4\\
            & DualGCN~\cite{zhuang2018dual} & 83.5 & 72.6 & 80.0 \\
            & JKNet~\cite{xu2018representation}\tnote{$\star$} & 81.1  &  69.8  & 78.1 \\
			& LGCN~\cite{gao2018large} &83.3 $\pm$ 0.5 & 73.0 $\pm$ 0.6 & 79.5 $\pm$ 0.2\\
			& Diff-GCN~\cite{klicpera2019diffusion} &  83.6 $\pm$ 0.2 & 73.4 $\pm$ 0.3 & 79.6 $\pm$ 0.4\\
			& APPNP~\cite{klicpera2018predict} & 84.3 $\pm$ 0.2 & 71.1 $\pm$ 0.4 & 79.7 $\pm$ 0.3 \\
			& g-U-Nets~\cite{gao2019graph} & 84.4 $\pm$ 0.6 &  73.2 $\pm$ 0.5 & 79.6 $\pm$ 0.2 \\
			& GAT~\cite{velivckovic2017graph} & 83.0 $\pm$ 0.7 & 72.5 $\pm$ 0.7 &  79.0 $\pm$ 0.3\\
			\hline
			\multirow{3}{*}{\rotatebox{90}{\hspace*{+2pt} Abl.}} & No LayerNorm & 83.8 $\pm$ 0.6 & 71.1 $\pm$ 0.5  & 79.8 $\pm$ 0.2 \\
			& No Diffusion & 83.0 $\pm$ 0.4 & 71.6 $\pm$ 0.4 & 79.3 $\pm$ 0.3\\
			& No Feed-Forward\tnote{$\diamond$}  & 84.9 $\pm$ 0.4 & 72.2 $\pm$ 0.3 & 80.9 $\pm$ 0.3 \\
			& No (LayerNorm + Feed-Forward)  & 84.3 $\pm$ 0.6 & 72.6 $\pm$ 0.4 & 79.6 $\pm$ 0.4 \\
		    \hline
			& \textbf{\name} & \textbf{85.4} $\pm$ 0.6 &  \textbf{73.7} $\pm$ 0.5 & \textbf{81.4} $\pm$ 0.2 \\
			\hline
		\end{tabularx}
		\begin{tablenotes}
		\item [$\star$]: based on the implementation in https://github.com/DropEdge/DropEdge;
		\item [$\diamond$]: replace the feed forward layer with \textit{elu} used in GAT.
		\end{tablenotes}
	\end{threeparttable}
	}\caption{Node classification accuracy on Cora, Citeseer, Pubmed. \name achieves state-of-the-art.}
	\label{table:classificationResults}
\end{table}

\paragraph{Experimental Setup} For datasets Cora, Citeseer and Pubmed, we use 6 \name blocks with hidden dimension 512 and 8 attention heads. For the large-scale ogbn-arxiv dataset, we use 2 \name blocks with hidden dimension 128 and 8 attention heads. 
Refer to Appendix for detailed description of all hyper-parameters and evaluation settings.

\begin{table*}[!h]
	\centering
	\resizebox{1.85\columnwidth}{!}{%
	\begin{threeparttable}
		\begin{tabularx}{1.15\textwidth}{l c c c c c c c}
			\hline
			Data & \shortstack{GCN \\ \cite{kipf2016semi}} & \shortstack{GraphSAGE \\ \cite{hamilton2017inductive}}  & \shortstack{JKNet \\ \cite{xu2018representation}} &
			\shortstack{DAGNN \\ \cite{liu2020towards}} &
			\shortstack{GaAN \\ \cite{zhang2018gaan}} & \textbf{\name}  \\
			\hline
			ogbn-arxiv  & 71.74 $\pm$ 0.29 & 71.49 $\pm$ 0.27 & 72.19 $\pm$ 0.21 & 72.09 $\pm$ 0.25 & 71.97 $\pm$ 0.24 & \textbf{72.76} $\pm$ 0.14\\
			\hline
		\end{tabularx}
	\end{threeparttable}
	}
	\caption{Node classification accuracy on the OGB Arxiv dataset.}
	\label{table:OGBclassificationResults}
\end{table*}

\paragraph{Results} 
\name achieves the best on all datasets (Tables \ref{table:classificationResults} and \ref{table:OGBclassificationResults})~\footnote{\noindent We also compared to GAT and Diffusion-GCN (with LayerNorm and feed-forward Layer) over random splits in Appendix.}, out-performing mutlihop baselines such as Diffusion GCN, APPNP and JKNet.
The baseline performance and their embedding dimensions are from the previous papers. Appendix Table 6 further demonstrates that large 512 dimension embedding only benefits the expressive \name, whereas GAT and Diffusion GCN performance degrades.

\paragraph{Ablation study} We report (Table \ref{table:classificationResults}) the model performance after removing each component of \name (layer normalization, attention diffusion and feed forward layers) from every \name layer. Note that the model is equivalent to GAT without these three components. We observe that diffusion and layer normalization play a crucial role in improving the node classification performance for all datasets. 
Since \name computes attention diffusion in a recursive manner, layer normalization is crucial in ensuring training stability~\cite{ba2016layer}. Meanwhile,
comparing to GAT (see the next-to-last row of Table~\ref{table:classificationResults}), attention diffusion allows multi-hop attention in every layer to benefit node classification.

\subsection{Task 2: Knowledge Graph Completion}

\paragraph{Datasets} We evaluate \name on standard benchmark knowledge graphs: WN18RR~\cite{dettmers2018convolutional} and FB15K-237~\cite{toutanova2015observed}. See the statistics of these KGs in Appendix.

\paragraph{Baselines} We compare \name with state-of-the-art baselines, including (1) translational distance based models: TransE~\cite{bordes2013translating} and its latest extension RotatE~\cite{sun2019rotate}, OTE~\cite{tang2019orthogonal} and ROTH~\cite{chami2020low}; (2) semantic matching based models: ComplEx~\cite{trouillon2016complex},
QuatE~\cite{zhang2019quaternion},
CoKE~\cite{wang2019CoKE}, ConvE~\cite{dettmers2018convolutional}, DistMult~\cite{yang2014embedding}, TuckER~\cite{balazevic2019tucker} and AutoSF~\cite{zhang2020autosf}; (3) GNN-based models: R-GCN~\cite{schlichtkrull2018modeling}, SACN~\cite{shang2019end} and A2N~\cite{bansal2019a2n}.

\paragraph{Experimental Setup} We use the multi-layer \name as encoder for both FB15k-237 and WN18RR. We randomly initialize the entity embedding and relation embedding as the input of the encoders, and set the dimensionality of the initialized entity/relation vector as 100 used in DistMult~\cite{yang2014embedding}. We select other \name model hype-parameters, including number of layers, hidden dimension, head number, top-$k$, learning rate, hop number, teleport probability $\alpha$ and dropout ratios (see the settings of these parameter in Appendix), by a random search during the training.   

\paragraph{Training procedure} 
We use the standard training procedure used in previous KG embedding models~\cite{balazevic2019tucker,dettmers2018convolutional} (Appendix for details).
We follow an encoder-decoder framework: The encoder applies the proposed \name model to compute the entity embeddings. 
The decoder makes link prediction given the embeddings. 
To show the power of \name, we employ a simple decoder DistMult~\cite{yang2014embedding}.

\begin{table*}[t]
\small
	\centering
	\resizebox{1.82\columnwidth}{!}{%
	\begin{threeparttable}
		\begin{tabularx}{0.92\textwidth}{l l l l l l| l l l l l} 
			\hline
			\multirow{2}{*}{\textbf{Models}} & \multicolumn{5}{c|}{\textbf{WN18RR}} & \multicolumn{5}{c}{\textbf{FB15k-237}}\\
			\cline{2-11}
			 & MR & MRR & H@1 & H@3 & H@10 & MR & MRR & H@1 & H@3 & H@10\\
			\hline
			TransE~\cite{bordes2013translating} & 3384 & .226 & - & - & .501 & 357 & .294 & - & - & .465\\
			RotatE~\cite{sun2019rotate} & 3340 & .476 & .428 & .492 & .571 & 177 & .338 & .241 & .375 & .533\\
			OTE~\cite{tang2019orthogonal} & - & .491 & .442 & .511 & .583 & - & .361 & .267 & .396 & .550\\
			ROTH~\cite{chami2020low} & - & .496 & .449 & .514 & .586 & - & .344 & .246 & .380 & .535\\
			\hline
			ComplEx~\cite{trouillon2016complex} & 5261 & .44 & .41 & .46 & .51 & 339 & .247 & .158 & .275 & .428\\
 			QuatE~\cite{zhang2019quaternion} & 2314 & .488 & .438 & .508 & .582 & - & .366 & .271 & .401 & .556\\
 			CoKE~\cite{wang2019CoKE} & - & .475  & .437 & .490 & .552 & - & .361 & .269 & .398 & .547 \\
 			ConvE~\cite{dettmers2018convolutional} & 4187 & .43 & .40 & .44 & .52 & 244 & .325 & .237 & .356 & .501\\
 			DistMult~\cite{yang2014embedding} & 5110 & .43 & .39 & .44 & .49 & 254 & .241 & .155 & .263 & .419\\
			TuckER~\cite{balazevic2019tucker} & - & .470 & .443 & .482 & .526 & - & .358 & .266 & .392 & .544  \\
			AutoSF~\cite{zhang2020autosf} & - & .490 & .451 & - & .567 & - & .360 & .267 & - & .552  \\
			\hline
			R-GCN~\cite{schlichtkrull2018modeling} & - & - & - & - & - & - & .249 & .151 & .264 & .417\\
			SACN~\cite{shang2019end} & - & .47 & .43 & .48 & .54 & - & .35 & .26 & .39 & .54\\
			A2N~\cite{bansal2019a2n} & - & .45 & .42 & .46 & .51 & - & .317 & .232 & .348 & .486  \\
			\hline
{\bf \name + DistMult} & {2545} & {.502} & {.459} & {.519}  & {.589}  & {138}  & {.369} & {.275}  & {.409}  & {.563} \\
\hline
		\end{tabularx}
	\end{threeparttable}
	}\caption{KG Completion on WN18RR and FB15k-237. \name achieves state of the art.}
	\label{table:kgresults}
\end{table*}

\begin{figure*}[!t]
    \centering
    \begin{subfigure}[b]{0.25\textwidth}
        \includegraphics[width=0.85\textwidth]{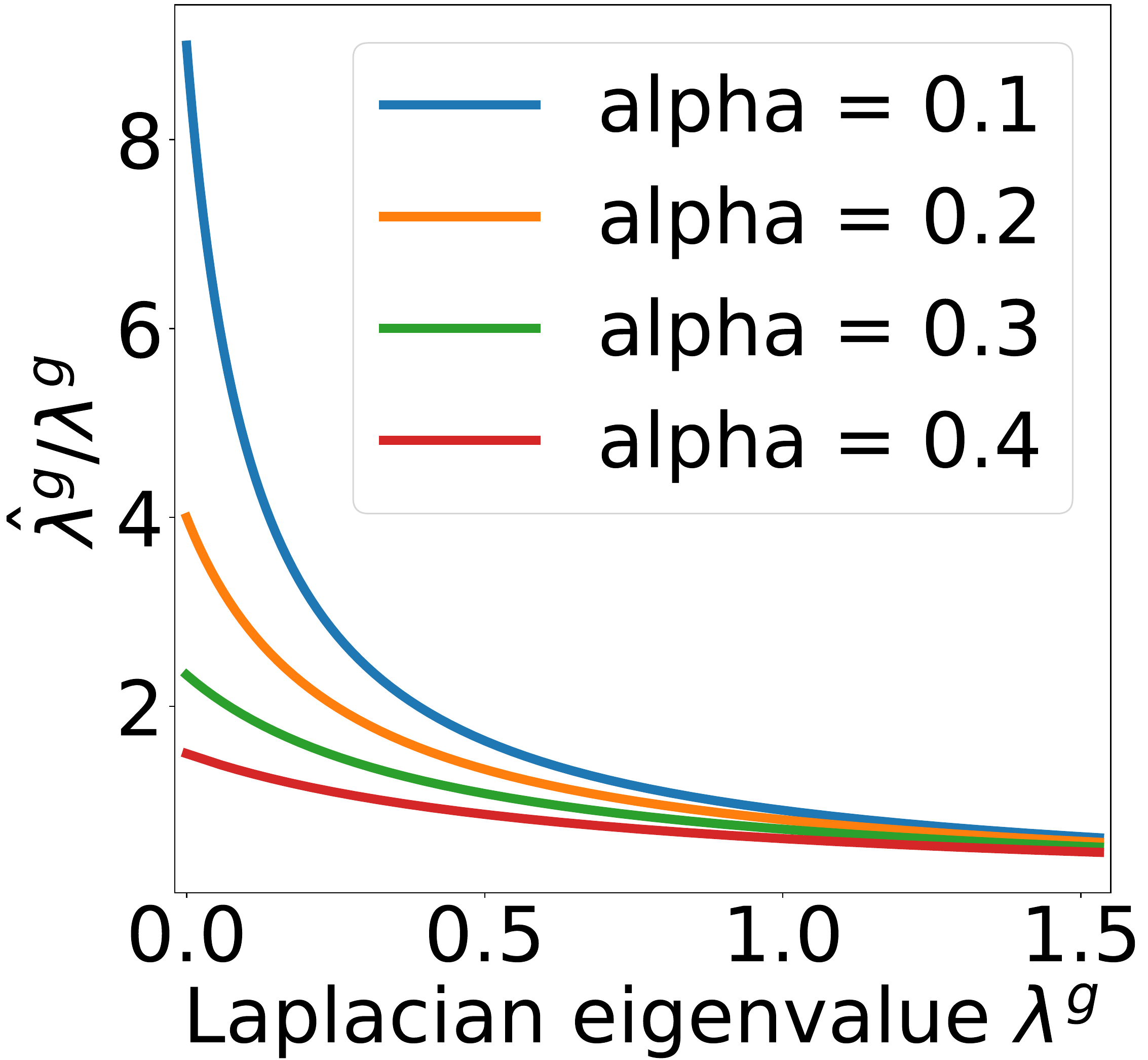}
    \end{subfigure}
    \begin{subfigure}[b]{0.25\linewidth}
          \includegraphics[width=0.85\textwidth]{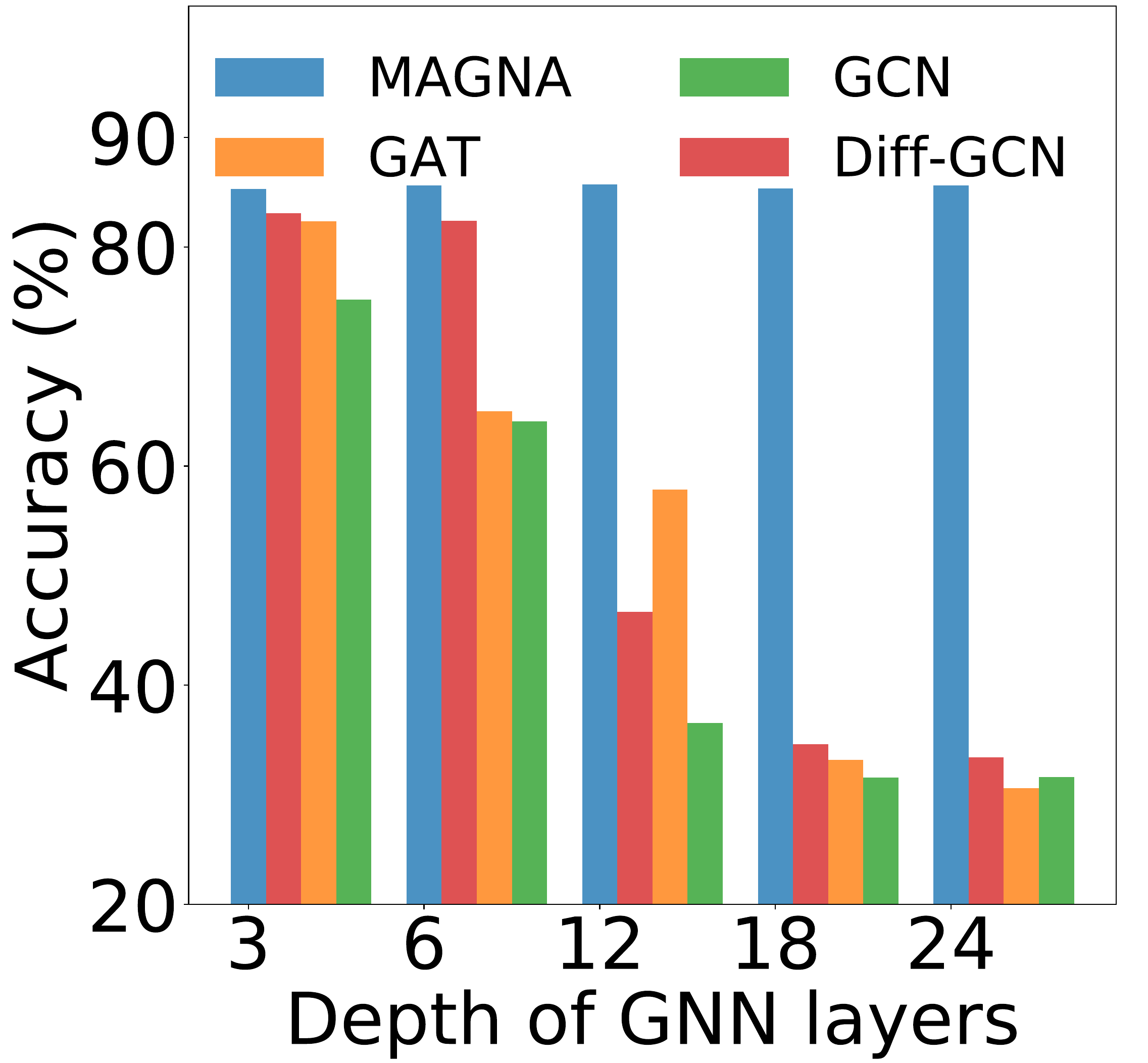}
    \end{subfigure}%
    \begin{subfigure}[b]{0.24\textwidth}
        \includegraphics[width=0.85\textwidth]{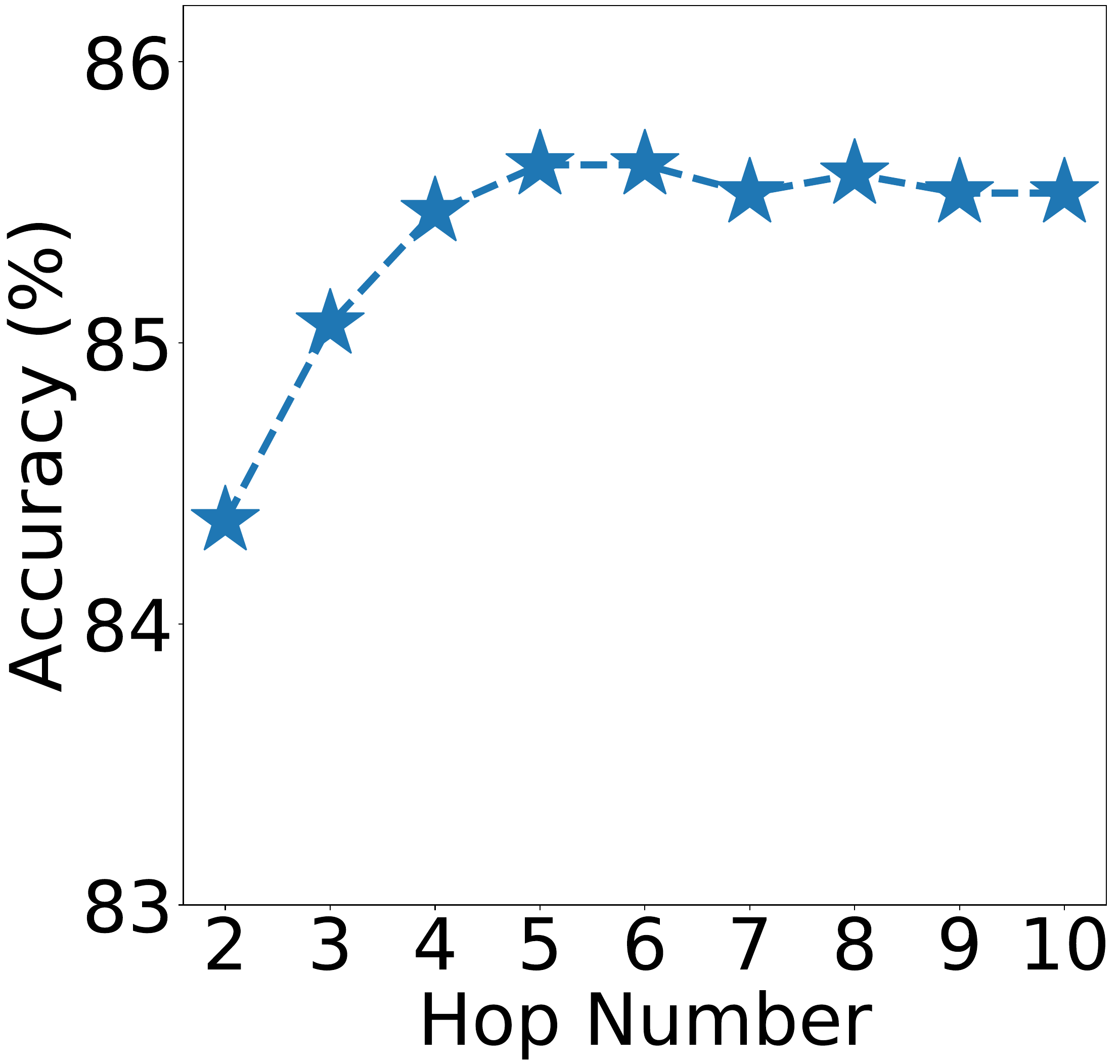}
    \end{subfigure}
    \begin{subfigure}[b]{0.24\textwidth}
        \includegraphics[width=0.85\textwidth]{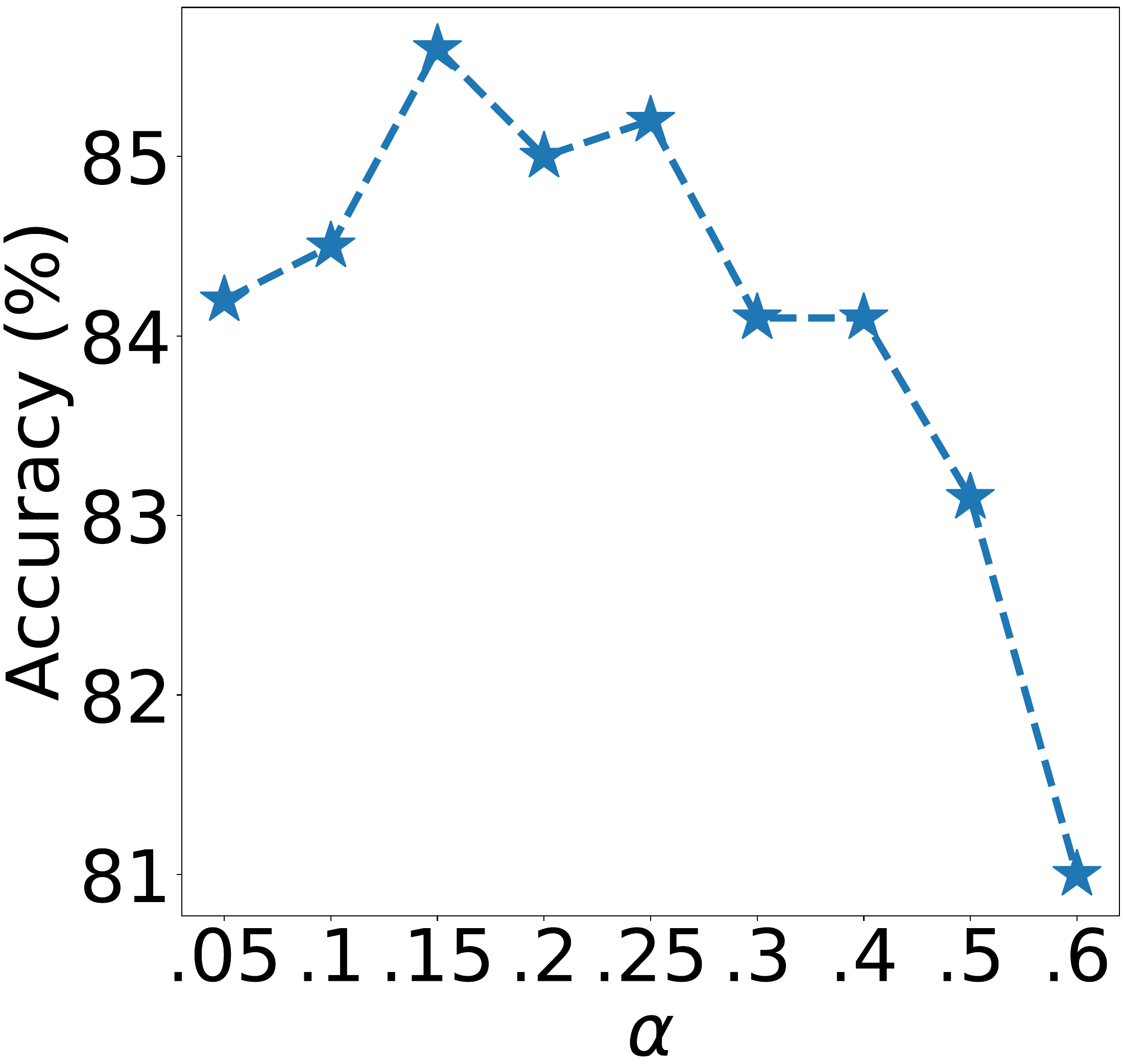}
    \end{subfigure}
    \caption{Analysis of \name on Cora. (a) Influence of \name on Laplacian eigenvalues. (b) Effect of depth on performance. (c) Effect of hop number $K$ on performance. (d) Effect of teleport probability $\alpha$. }
    \label{fig:analysis}
\end{figure*}

\paragraph{Evaluation} We use the standard split for the benchmarks, and the standard testing procedure of predicting tail (head) entity given the head (tail) entity and relation type.
We exactly follow the evaluation used by all previous works, namely the Mean Reciprocal Rank (MRR), Mean Rank (MR), and hit rate at $K$ (H@K). 
See Appendix for a detailed description of this standard setup.

\paragraph{Results} \name achieves new state-of-the-art in knowledge graph completion on all four metrics (Table~\ref{table:kgresults}).
\name compares favourably to both the most recent shallow embedding methods (QuatE), and deep embedding methods (SACN). Note that with the same decoder (DistMult), \name using its own embeddings achieves drastic improvements over using the corresponding DistMult embeddings.

\subsection{\name Model Analysis}
\label{sec:analysis}
Here we present (1) spectral analysis results, (2) robustness to hyper-parameter changes, and (3) attention distribution analysis to show the strengths of \name.


\paragraph{Spectral Analysis: Why \name works for node classification?}
We compute the eigenvalues of the graph Laplacian of the attention matrix $\mathbf{A}$, $\hat{\lambda}_{i}^{g}$, and compare to that of the diffused matrix $\mathcal{A}$, $\lambda_{i}^{g}$. Figure~\ref{fig:analysis} (a) shows the ratio $\hat{\lambda}_{i}^{g}/\lambda_{i}^{g}$ on the Cora dataset. Low eigenvalues corresponding to large-scale structure in the graph are amplified (up to a factor of 8), while high eigenvalues corresponding to eigenvectors with noisy information are suppressed \cite{klicpera2019diffusion}.

\paragraph{\name Model Depth} 
Here we conduct experiments by varying the number of GCN, Diffusion-GCN (PPR based) GAT and our \name layers to be 3, 6, 12, 18 and 24 for node classification on Cora. Results in Fig. ~\ref{fig:analysis} (b) show that deep GCN, Diffusion-GCN and GAT (even with residual connection) suffer from degrading performance, due to the over-smoothing problem~\cite{li2018deeper,wang2019improving}. In contrast, the \name model achieves consistent best results even with 18 layers, making deep \name model robust and expressive. Notice that GAT with 18 layers cannot out-perform \name with 3 layers and $K$=6 hops, although they have the same receptive field.

\paragraph{Effect of $K$ and $\alpha$} Figs.~\ref{fig:analysis} (c) and (d) report the effect of hop number $K$ and teleport probability $\alpha$ on model performance. We observe significant increase in performance when considering multi-hop neighbors information ($K>1$). However, increasing the hop number $K$ has a diminishing returns, for $K \geq 6$. Moreover, we find that the optimal $K$ is correlated with the largest node average shortest path distance (e.g., 5.27 for Cora). This provides a guideline for choosing the best $K$.

We also observe that the accuracy drops significantly for larger $\alpha > 0.25$. This is because small $\alpha$ increases the low-pass effect (Fig. \ref{fig:analysis} (a)). 
However, $\alpha$ being too small causes the model to only focus on the most large-scale graph structure and have lower performance.

\begin{figure}[!t]
    \centering
    \includegraphics[width=0.28\textwidth]{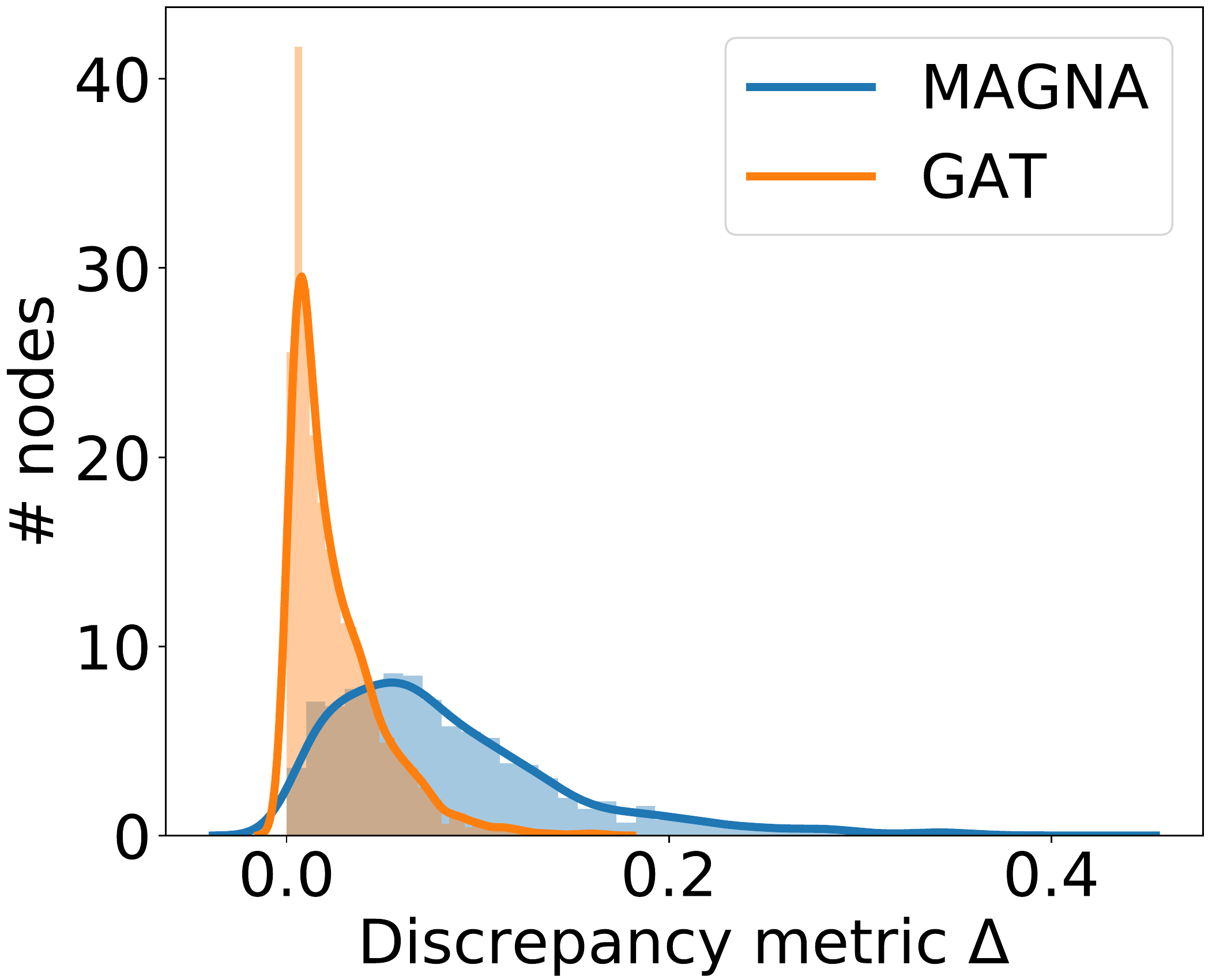} 
    \caption{Attention weight distribution on Cora.}
    \label{fig:attentionDistribution}
\end{figure}
\paragraph{Attention Distribution} Last we also analyze the learned attention scores of GAT and \name.
We first define a discrepancy metric over the attention matrix $\mathbold{A}$ for node $v_i$ as $\Delta_{i} =\frac{\parallel \mathbold{A}_{[i,:]} - U_{i}\parallel}{\text{degree}(v_i)}$~\cite{shanthamallu2020regularized}, where $U_i$ is the uniform distribution score for the node $v_{i}$. $\Delta_{i}$ gives a measure of how much the learnt attention deviates from an uninformative uniform  distribution. Large  $\Delta_{i}$ indicates more meaningful attention scores.
Fig.~\ref{fig:attentionDistribution} shows the distribution of the discrepancy metric of the attention matrix of the 1st head w.r.t. the first layer of \name and GAT. Observe that attention scores learned in \name have much larger discrepancy. This shows that \name is more powerful than GAT in distinguishing important nodes and assigns attention scores accordingly.

\section{Related Work}
\cut{
\jure{This related work is a bit too scattered. I suggest we really focus it on GNNs that use attention and on any other works that use PPR and self-attention. Are there any other works that claim to have transformer on a graph? If they are we need to discuss them here.
\\
We can skip the discussion of KGs, transformers, GNNs, etc. Let's just focus on attention mechanisms for GNNs and anyone else who tries to do self-attention (see the PPR paper I sent you a few days ago).}
}

Our proposed \name belongs to the family of Graph Neural Network (GNN) models~\cite{battaglia2018relational,wu2019comprehensive,kipf2016semi,hamilton2017inductive}, while taking advantage of graph attention and diffusion techniques.


\paragraph{Graph Attention Neural Networks (GATs)} generalize attention operation to graph data. GATs allow for assigning different importance to nodes of the same neighborhood at the feature aggregation step~\cite{velivckovic2017graph}. Based on such framework, different attention-based GNNs have been proposed, including GaAN~\cite{zhang2018gaan}, AGNN~\cite{thekumparampil2018attention}, GeniePath~\cite{liu2019geniepath}. However, these models only consider direct neighbors for each layer of feature aggregation, and suffer from over-smoothing when they go deep~\cite{wang2019improving}. 

\paragraph{Diffusion based Graph Neural Network} Recently Graph Diffusion Convolution~\cite{klicpera2019diffusion,klicpera2018predict} proposes to aggregate information from a larger (multi-hop) neighborhood at each layer, by sparsifying a generalized form of graph diffusion. 
This idea was also explored in~\cite{Liao2019ICLR,Luan2019NIPS,continuous2019,klicpera2018predict} for multi-scale Graph Convolutional Networks.
However, these methods do not incorporate attention mechanism which is crucial to model performance, and do not make use of edge embeddings (e.g., Knowledge graph)~\cite{klicpera2019diffusion}. 
Our approach defines a novel multi-hop context-dependent self-attention GNN which resolves the over-smoothing issue of GAT architectures~\cite{wang2019improving}. 
\cite{isufi2020edgenets,cucurull2018convolutional,feng2019attention} also extends attention mechanism for multi-hop information aggregation, but they require different set of parameters to compute the attention to neighbors of different hops, making these approaches much more expensive compared to MAGNA, and were not extended to the knowledge graph settings.

\section{Conclusion}
We proposed \namelong (\name), which brings together benefits of graph attention and diffusion techniques in a single layer through attention diffusion, layer normalization and deep aggregation. 
\name enables context-dependent attention between any pair of nodes in the graph in a single layer, enhances large-scale structural information, and learns more informative attention distribution.
\name improves over all state-of-the-art methods on the standard tasks of node classification and knowledge graph completion.

{
\bibliographystyle{named}
\bibliography{ijcai21}
}

\clearpage
\newpage
\appendix
\renewcommand{\arraystretch}{1.04}

\section{Attention Diffusion Approximation Proposition}
As mentioned in Section 2.2, we use the following equation and proposition to efficiently approximate the attention diffused feature aggregation $\mathcal{A} \mathbold{H}^{(l)}$.
\begin{equation}\label{eq:approxGAT_appendix}
    \begin{aligned}
    & \mathbold{Z}^{(0)} \quad = \mathbold{H}^{(l)} \\
    & \mathbold{Z}^{(k+1)} = (1 - \alpha)\mathbold{A}\mathbold{Z}^{(k)}+ \alpha\mathbold{Z}^{(0)}\\
    \end{aligned}
\end{equation}

\begin{prop}\label{prop:approxGAT_appendix}
    $\lim_{K\rightarrow\infty}Z^{(K)} = \mathcal{A}\mathbold{H}^{(l)}$ 
\end{prop}

\begin{proof}
Let $K > 0$ be the total number of iterations (i.e., hop number of graph attention diffusion) and we approximate $\hat{\mathbold{H}}^{(l)}$ by $\mathbold{Z}^{(K)}$. 
After $K$-th iteration, we can get
\begin{equation}\label{eq:approxGATFinal}
    \mathbold{Z}^{K} = ((1 - \alpha)^{K}\mathbold{A}^{K} + \alpha\sum_{i=0}^{K-1}(1-\alpha)^{i}\mathbold{A}^{i})\mathbold{H}^{(l)}
\end{equation}
The term $(1 - \alpha)^{K}\mathbold{A}^{K}$ converges to 0 as $\alpha \in (0,1]$ and $\mathbold{A}^{K}_{i,j} \in (0,1]$ when $K \rightarrow \infty$, and thus $\lim_{K\rightarrow\infty}Z^{(K)} = (\sum_{i=0}^{\infty}\alpha(1-\alpha)^{i}\mathbold{A}^{i})\mathbold{H}^{(l)}$ = $\mathcal{A}\mathbold{H}^{(l)}$.
\end{proof}

\xhdr{Complexity analysis} Suppose the graph contains $E$ edges, using the above approximation, there
are $O(|E|)$ message communications in total, with a constant factor corresponding to the number of hops $K$. In practice, we find that choosing the values of $K$ such that $3 \leq K \leq 10$ results in good model performance.

\section{Connection to Transformer}
\label{app:connection_transformer}
Given a sequence of tokens, the Transformer architecture makes uses of multi-head attention between all pairs of tokens, and can be viewed as performing message-passing on a fully connected graph between all tokens.
A na\"ive application of Transformer on graphs would require computation of all pairwise attention values. Such approach, however, would not make effective use of the graph structure, and could not scale to large graphs.
In contrast, Graph Attention Network~\cite{velivckovic2017graph} leverages the graph structure and only computes attention values and perform message passing between direct neighbors. However, it has a limited receptive field (restricted to one-hop neighborhood) and the attention score (Figure 1) that is independent of the multi-hop context for prediction. 

Transformer consists of self-attention layer followed by feed-forward layer. We can organize the self-attention layer in transformer as the following:
\begin{equation}\label{eq:transformer}
    \text{Attention}(\mathbold{Q}, \mathbold{K}, \mathbold{V}) = \underbrace{\text{softmax}(\frac{\mathbold{Q}\mathbold{K}^{T}}{\sqrt{d}})}_\textrm{Attention matrix}\mathbold{V}
\end{equation}
where $\mathbold{Q} = \mathbold{K} = \mathbold{V}$. The $\text{softmax}$ part can be demonstrated as an attention matrix computed by scaled dot-product attention over a complete graph\footnote{All nodes are connected with each other.} with self-loop.
Computation of attention over complete graph is expensive, Transformers are usually limited by a fixed-length context (e.g., 512 in BERT~\cite{devlin2019bert}) in the setting of language modeling, and thus cannot handle large graphs.
Therefore direct application of the transformer model cannot capture the graph structure in a scalable way. 

In the past, graph structure is usually encoded implicitly by special position embeddings~\cite{zhang2020graph} or well-designed attention computation~\cite{wang2019tree,nguyen2020tree}. However, none of the methods can compute attention between any pair of nodes at each layer. 

In contrast, essentially \name places a prior over the attention values via Personalized PageRank, allowing it to compute the attention between any pair of two nodes via attention diffusion, without any impact on its scalability.
In particular, \name can handle large graphs as they are usually quite sparse and the graph diameter is usually quite smaller than graph size in practice, resulting in very efficient attention diffusion computation.

\section{Spectral Analysis Background and Proof for Proposition 2}
\xhdr{Graph Fourier Transform} Suppose $\mathbold{A}_{N\times N}$ represents the attention matrix of graph $\mathcal{G}$ with $\mathbold{A}_{i,j} \geq 0$, and $\sum_{j=1}^{N}\mathbold{A}_{i,j} = 1$. Let $\mathbold{A} = \mathbold{V}\mathbold{\Lambda} \mathbold{V}^{-1}$ be the Jordan’s decomposition of graph attention matrix $\mathbold{A}$, where $\mathbold{V}$ is the square $N \times N$ matrix whose $i$-th column is the eigenvector $\mathbold{v}_i$ of $\mathbold{A}$, and $\mathbold{\Lambda}$ is the diagonal matrix whose diagonal elements are the corresponding eigenvalues, i.e., $\mathbold{\Lambda}_{i,i} = \lambda_i$. Then, for a given vector $\mathbold{x}$, its \emph{Graph Fourier Transform} \cite{sandryhaila2013discrete} is defined as 
\begin{equation}
    \mathbold{\hat{x}} = \mathbold{V}^{-1}\mathbold{x},
\end{equation}
where $\mathbold{V}^{-1}$ is denoted as graph Fourier transform matrix. The \emph{Inverse Graph Fourier
Transform} is defined as $\mathbold{x} = \mathbold{V}\mathbold{\hat{x}}$, which reconstructs the signal from its spectrum. Based on the graph Fourier transform, we can define a graph convolution operation on $\mathcal{G}$ as $\mathbold{x}\otimes_{\mathcal{G}}\mathbold{y}$ = $\mathbold{V}(\mathbold{V}^{-1}\mathbold{x}\odot\mathbold{V}^{-1}\mathbold{y})$, where $\odot$ denotes the
element-wise product.

\xhdr{Graph Attention Diffusion Acts as A Polynomial Graph Filter} A graph filter~\cite{tremblay2018design,sandryhaila2013discrete,sandryhaila2013discrete2} $\mathbold{h}$ acts on $\mathbold{x}$ as $\mathbold{h}(A)\mathbold{x}$ = $\mathbold{V}\mathbold{h}(\Lambda)\mathbold{V}^{-1}$, where $\mathbold{h}(\Lambda) = diag(\mathbold{h}(\lambda_1) \cdots \mathbold{h}(\lambda_N))$. A common choice for $\mathbold{h}$ in the literature is a polynomial
filter of order $M$, since it is linear and
shift invariant \cite{sandryhaila2013discrete,sandryhaila2013discrete2}.
\begin{equation}
    \mathbold{h}(\mathbold{A}) = \sum_{\ell=0}^{M}\beta_{\ell} \mathbold{A}^{\ell} = \sum_{\ell=0}^{M}\beta_{\ell}(\mathbold{V}\mathbold{\Lambda} \mathbold{V}^{-1})^{\ell} = \mathbold{V}(\sum_{\ell=0}^{M}\beta_{\ell}\Lambda^{\ell})\mathbold{V}^{-1}.
\end{equation}
Comparing to the graph attention diffusion $\mathcal{A} = \sum_{i=0}^{\infty}\alpha(1-\alpha)^{i}\mathbold{A}^{i}$, if we set $\beta_{\ell} = \alpha(1-\alpha)^{\ell}$, we can view \textbf{graph attention diffusion as a polynomial filter}.

\xhdr{Spectral Analysis} 
The eigenvectors of the power matrix $\mathbold{A}^{2}$ are same as $\mathbold{A}$, since $\mathbold{A}^{2}$ = $(\mathbold{V}\mathbold{\Lambda}\mathbold{V}^{-1}) (\mathbold{V}\mathbold{\Lambda}\mathbold{V}^{-1})$ = $\mathbold{V}\mathbold{\Lambda}(\mathbold{V}^{-1} \mathbold{V})\mathbold{\Lambda}\mathbold{V}^{-1}$ = $\mathbold{V}\mathbold{\Lambda}^{2}\mathbold{V}^{-1}$. By that analogy, we can get that $\mathbold{A}^{n} = \mathbold{V}\mathbold{\Lambda}^{n}\mathbold{V}^{-1}$. Therefore, the summation of the power series of $\mathbold{A}$ has the same eigenvectors as $\mathbold{A}$. 
Therefore by properties of eigenvectors and Equation \ref{eq:attentionDiffusion}, we obtain:
\begin{prop}\label{prop:eigenvector}
The set of eigenvectors for $\mathcal{A}$ and $\mathbold{A}$ are the same.
\end{prop}

\begin{lemma}\label{lemma:spectralAnaGDT}  Let $\lambda_{i}$ and $\hat{\lambda}_{i}$ be the $i$-th eigenvalues of $\mathbold{A}$ and $\mathcal{A}$, respectively. Then, we have
\begin{equation}
    \hat{\lambda}_{i} = \sum_{\ell=0}^{\infty}\beta_{\ell} \lambda_{i}^{\ell} = \sum_{\ell=1}^{\infty}\alpha(1-\alpha)^{\ell} \lambda_{i}^{\ell} = \frac{\alpha}{1 - (1-\alpha)\lambda_{i}}
\end{equation}
\end{lemma}

\begin{proof}
The symmetric normalized graph Laplacian of $\mathcal{G}$ is $\mathbold{L}_{sym} = \mathbold{I} - \mathbold{D}^{-\frac{1}{2}}\mathbold{A}\mathbold{D}^{-\frac{1}{2}}$, where $\mathbold{D}$ = $diag([d_1, d_2, \cdots, d_N])$, and $d_i = \sum_{j=1}^{\mathbold{A}_{i,j}}$.
As $\mathbold{A}$ is the attention matrix of graph $\mathcal{G}$, $d_i = 1$ and thus $\mathbold{D} = \mathbold{I}$. Therefore, $\mathbold{L}_{sym}$ = $\mathbold{I} - \mathbold{A}$. Let $\lambda_i$ be the eigenvalues of $\mathbold{A}$, the eigenvalues of the symmetric normalized Laplacian of $\mathcal{G}$ $\mathbold{L}_{sym}$ is $\bar{\lambda}_i = 1 - \lambda_i$. Meanwhile, for every eigenvalue $\bar{\lambda}_i$ of the normalized graph
Laplacian $\mathbold{L}_{sym}$, we have $0 \leq \bar{\lambda}_i \leq 2$~\cite{mohar1991laplacian}, and thus $-1 \leq  \lambda_i \leq 1$. As $0 < \alpha < 1$ and thus $|(1 -\alpha)\lambda_{i}| \leq (1 -\alpha) < 1$. Therefore, $((1 -\alpha)\lambda_{i})^{K} \rightarrow 0$ when $K \rightarrow \infty$, and $\hat{\lambda}_{i} = \lim_{K\rightarrow\infty}\sum_{\ell=0}^{K}\alpha(1-\alpha)^{\ell} \lambda_{i}^{\ell} = \lim_{K\rightarrow\infty}\frac{\alpha(1 - ((1 - \alpha)\lambda_{i})^{K})}{1 - (1-\alpha)\lambda_{i}} = \frac{\alpha}{1 - (1-\alpha)\lambda_{i}}$.
\end{proof}

Section 3 further defines the eigenvalues of the Laplacian matrices, $\hat{\lambda}_{i}^{g}$ and $\lambda_{i}^{g}$ respectively.
They satisfy: $\hat{\lambda}_{i}^{g} = 1 - \hat{\lambda}_{i}$ and $\lambda_{i}^{g} = 1 - \lambda_{i}$, and $\lambda_{i}^{g} \in [0,2]$ (proved by ~\cite{ng2002spectral}).

\section{Graph Learning Tasks}

Node classification and knowledge graph link prediction are two representative and  common tasks in graph learning. 
We first define the task of node classification:

\begin{definition}
\label{def:nodeclassification}
    \textbf{Node classification} Suppose that $\mathbold{X} \in \mathbb{R}^{N_{n}\times d}$ represents the node input features, where each row $\mathbold{x}_{i}$ = $\mathbold{X}_{i:}$ is a $d$-dimensional vector of attribute values of node $v_i \in \mathcal{V}$ ($1\leq i \leq N$). $\mathcal{V}_{l} \subset \mathcal{V}$ consists of a set of labeled nodes, and the labels are from $\mathcal{T}$, node classification is to learn the map function $\mathit{f}: (\mathbold{X}, \mathcal{G}) \rightarrow \mathcal{T}$, which predicts the labels of the remaining un-labeled nodes $\mathcal{V}/\mathcal{V}_{l}$.
\end{definition}

\noindent\textbf{Knowledge graph (KG)} is a heterogeneous graph describing entities and their typed relations to each other. KG is defined by a set of entities (nodes) $v_i \in \mathcal{V}$, and 
a set of relations (edges) $e = (v_i, r_k, v_j)$ connecting nodes $v_i$ and $v_j$ via relation $r_k$. 
We then define the task of knowledge graph completion:

\begin{definition}\label{def:linkPrediction}
    \textbf{KG completion} refers to the task of predicting an entity that has a specific relation with another given entity~\cite{bordes2013translating}, i.e., predicting head $h$ given a pair of relation and entity $(r, t)$ or predicting tail $t$ given a pair of head and relation $(h, r)$.
\end{definition}

\section{Dataset Statistics}
\xhdr{Node classification} We show the dataset statistics of the node classification benchmark datasets in Table \ref{table:nodeclass_dataset}.

\begin{table*}[!ht]
	\centering
	\footnotesize
	\caption{Statistical Information on Node Classification Benchmarks}
	\begin{threeparttable}
		\begin{tabularx}{0.625\textwidth}{l c c c c c} 
			\hline
			Name & Nodes & Edges & Classes & Features  & Train/Dev/Test \\
			\hline
			Cora & 2,708 & 5,429 & 7 & 1,433  & 140/500/1,000  \\
			\hline
			Citeseer & 3,327 & 4,732 & 6& 3,703 & 120/500/1,000  \\
			\hline
			Pubmed & 19,717 & 88,651 & 3 & 500 &  60/500/1,000 \\
			\hline
			ogbn-arxiv\tnote{$\dagger$} & 169,343 & 1,166,243 & 40 & 128 & 90,941/29,799/48,603 \\
			\hline
		\end{tabularx}
		\begin{tablenotes}
        \item [$\dagger$] The data is available at \url{https://ogb.stanford.edu/docs/nodeprop/}.
	    \end{tablenotes}
	\end{threeparttable}
	\label{table:nodeclass_dataset}
\end{table*}

\xhdr{Knowledge Graph Link Prediction} We show the dataset statistics of the knowledge graph benchmarks in Table \ref{table:kgdataset}.
\begin{table*}[!h]
	\centering
	\footnotesize
	\caption{Statistical Information on KG Benchmarks}
	\begin{threeparttable}
		\begin{tabularx}{.645\textwidth}{r c c c c c c} 
			\hline
			{\textbf{Dataset}} & \#\textbf{Entities} & {\#\textbf{Relations}} & {\#\textbf{Train}} &  {\#\textbf{Dev}} & {\#\textbf{Test}} & {\#\textbf{Avg. Degree}} \\
			\hline
			WN18RR & 40,943 & 11 & 86,835 & 3034 & 3134 & 2.19 \\
			\hline
			FB15k-237 &14,541& 237& 272,115 & 17,535 & 20,466 & 18.17 \\
			\hline
		\end{tabularx}
	\end{threeparttable}
	\label{table:kgdataset}
\end{table*}

\section{Knowledge Graph Training and Evaluation}
\xhdr{Training}
The standard knowledge graph completion task training procedure is as follows.
We add the reverse-direction triple $(t, r^{-1}, h)$ for each triple $(h, r, t)$ to construct an undirected knowledge graph $\mathcal{G}$. 
Following the training procedure introduced in~\cite{balazevic2019tucker,dettmers2018convolutional}, we use 1-N scoring, i.e. we simultaneously
score entity-relation pairs $(h, r)$ and $(t, r^{-1})$ with all entities, respectively. We explore KL diversity loss with label smoothing as the optimization function. 

\xhdr{Inference time procedure}
For each test triplet $(h, r, t)$, the head $h$ is removed and replaced by each of the entities appearing in KG. Afterward, we remove from the corrupted triplets all the ones that appear either in the training, validation or test set. Finally, we score these corrupted triplets by the link prediction models and then sorted by descending order; the rank of $(h, r, t)$ is finally scored. This whole procedure is repeated while removing the tail $t$ instead of $h$. And averaged metrics are reported. We report mean reciprocal rank (MRR), mean rank (MR) and the proportion of correct triplets in the top $K$ ranks (Hits@$K$) for $K$ = 1, 3 and 10. Lower values of MR and larger values of MRR and Hits@$K$ mean better performance.


\section{Results}



\xhdr{Comparison to Diffusion GCN} Diffusion-GCN~\cite{klicpera2019diffusion} is also based on Personal Page-Rank (PPR) propagation. Specifically, it performs propagation over the adjacent matrix. Comparing to \name, there is no LayerNorm and Feed Forward layers in standard Diffusion GCN. To clarify how much of the gain in performance depends on the page-rank-based propagation compared to the attention propagation in \name, we add the two modules: LayerNorm and Feed Forward layer, into Diffusion-GCN, and conduct experiments over Cora, Citeseer and Pubmed, respectively. Table~\ref{table:diffGCN} shows the comparison results. From this table, we observe that adding layer normalization and feed forward layer gets the similar GNN structure to \name, but does not benefit for node classification. And this implies that the PPR based attention propagation in \name is more effective than PPR propagation over adjacent matrix in Diffusion GCN.


\begin{table}[t]
	\centering
	\resizebox{0.5\textwidth}{!}{%
	\begin{threeparttable}
		\begin{tabularx}{0.75\textwidth}{l| c c c} 
			\hline
			 Model & Cora & Citeseer & Pubmed\\
			\hline
			Diffusion-GCN (PPR)~\cite{klicpera2019diffusion} &  83.6 $\pm$ 0.2 & 73.4 $\pm$ 0.3 & 78.7 $\pm$ 0.4\tnote{$\star$}\\
			Diffusion-GCN (PPR) + LayerNorm + FF\tnote{$\diamond$} & 83.4 $\pm$ 0.4  & 72.3 $\pm$ 0.4  & 78.1$\pm$ 0.5 \\
			\hline
			GAT (hidden dimension = 512)\tnote{$\ast$} & 83.1 $\pm$ 0.5  & 71.3 $\pm$ 0.4  & 78.3$\pm$ 0.4 \\
			\hline
			\textbf{\name} & \textbf{85.4} $\pm$ 0.6 &  \textbf{73.7} $\pm$ 0.5 & \textbf{81.4} $\pm$ 0.2 \\
			\hline
		\end{tabularx}
		\begin{tablenotes}
		\item [$\diamond$]: FF: Feed Forward (FF) layer, and our implementation is based on Diffusion GCN in pytorch geometric (\url{https://github.com/rusty1s/pytorch_geometric}).
		\item [$\star$]: The best number derived from Diffusion GCN with ``PPR''. This is different from the number in Table 1 of main body, which comes from Diffusion GCN with ``Heat''.
		\item [$\ast$]: The results of GAT with the same hidden dimension in \name.
		\end{tablenotes}
	\end{threeparttable}
	}\caption{Comparison of Diffusion GCN (PPR), GAT (with hidden dimension = 512) and \name for Node classification accuracy on Cora, Citeseer, Pubmed.}
	\label{table:diffGCN}
\end{table}


\begin{table}[!h]
	\centering
	\resizebox{0.5\textwidth}{!}{%
	\begin{threeparttable}
		\begin{tabularx}{0.725\textwidth}{l | c c c} 
			\hline
			Model & Cora & Citeseer &  Pubmed\\
			\hline
			Diffusion GCN (PPR)~\cite{klicpera2019diffusion} & 83.6 $\pm$ 1.8 & 71.7 $\pm$ 1.1 & 79.6 $\pm$ 2.5\\
			Diffusion GCN (PPR) + (LayerNorm \& FF)  & 83.3 $\pm$ 1.9 & 69.7 $\pm$ 1.2 &  79.7 $\pm$ 2.3\\
			GAT~\cite{velivckovic2017graph} & 82.2 $\pm$ 1.4 & 71.0 $\pm$ 1.2 & 78.3 $\pm$ 3.0 \\
			GAT + (LayerNorm \& FF) & 82.5 $\pm$ 2.2 & 69.5 $\pm$ 1.1 & 78.2 $\pm$ 2.0 \\
			\hline
			\textbf{\name} & \textbf{83.6} $\pm$ 1.2 &  \textbf{72.1} $\pm$ 1.2 & \textbf{80.3} $\pm$ 2.4 \\
			\hline
		\end{tabularx}
	\end{threeparttable}
	}\caption{Results of \name on Citeseer, Cora and Pubmed with random splitting. The baselines are Diffusion GCN (PPR) and GAT.}
	\label{table:randomSplit}
\end{table}

\xhdr{Comparison over Random Splitting} We also randomly split nodes in each graph for training, validation and testing following the same ratio in the standard data splits.
We conduct experiments with \name, Diffusion GCN (PPR) and GAT over 10 different random splits over Cora, Citeseer and Pubmed, respectively. For each algorithm, we apply random search~\cite{bergstra2012random} for hype-parameter tuning. We report the average classification accuracy with standard deviation over 10 different splits in Table~\ref{table:randomSplit}. From this table, we observe that \name gets the best average accuracy over all data sets. However, it is noted that the standard deviation is quite large comparing to the that from the standard split in Table~\ref{table:diffGCN}. This means that the performance is quite sensitive to the graph split. This observation is consistent wit the conclusion in~\cite{hu2020ogb}: random splitting is often problematic in graph learning as they are impractical in real scenarios, and can lead to large performance variation. Therefore, in our main results, we use the standard split and also record standard deviation to demonstrate significance of the results.

\section{Hyper-parameter Settings} 

\noindent\textbf{Hyper-parameter settings for node classification.} The best models are selected according to the classification accuracy on the validation set by early stopping with window size 200.  

For each data set, the hyper-parameters are determined by a random search~\cite{bergstra2012random}, including learning rate, hop number, teleport probability $\alpha$ and dropout ratios. The hyper-parameter search space is show in Tables~\ref{table:randomsearch_nodeclass} (for Cora, Citeseer and Pubmed) and \ref{table:randomsearch_nodeclass_ogb} (for ogbn-arxiv).

\begin{table}[!h]
	\centering
	\footnotesize
	\begin{threeparttable}
		\begin{tabularx}{.5\textwidth}{l c c} 
			\hline
			Hyper-parameters & Search Space & Type\\
			\hline
			Hidden Dimension & 512 & Fixed\tnote{$\vdash$}\\
			Head Number & 8 & Fixed\\
			Layer Number & 6 & Fixed\\
			Learning rate & [$5\times 10^{-5}, 10^{-3}$] & Range\tnote{$\star$} \\ 
			Hop Number & [$2, 3, \cdots, 10$] & Choice\tnote{$\diamond$}\\
			Teleport probability $\alpha$ & [$0.05, 0.6$] & Range\\ 
			Dropout (attention, feature) & [$0.1, 0.6$] & Range\\
			Weight Decay & [$10^{-6}, 10^{-5}$] & Range\\
			Optimizer & Adam & Fixed\\
			\hline
		\end{tabularx}
		\begin{tablenotes}
		\item [$\vdash$] Fixed: a constant value; $\star$: Range: a value range with lower bound and higher bound; $\diamond$: Choice: a set of values.
	    \end{tablenotes}
	\end{threeparttable}
		\caption{Hyper-parameter search space used for node classification on Cora, Citeseer and Pubmed}
	\label{table:randomsearch_nodeclass}
\end{table}
\begin{table}[!th]
	\centering
	\footnotesize
	\begin{threeparttable}
		\begin{tabularx}{.45\textwidth}{l c c} 
			\hline
			Hyper-parameters & Search Space & Type\\
			\hline
			Hidden Dimension & 128 & Fixed\\
			Head Number & 8 & Fixed\\
			Layer Number & 2 & Fixed\\
			Learning rate & [$0.001, 0.01$] & Range \\ 
			Hop Number & [$3, 4, 5, 6$] & Choice\\
			Teleport probability $\alpha$ & [$0.05, 0.6$] & Range\\ 
			Dropout (attention, feature) & [$0.1, 0.6$] & Range\\
			Weight Decay & [$10^{-5}, 10^{-4}$] & Range\\
			Optimizer & Adam & Fixed\\
			\hline
		\end{tabularx}
	\end{threeparttable}
		\caption{Hyper-parameter search space for node classification on ogbn-arxiv}
	\label{table:randomsearch_nodeclass_ogb}
\end{table}

\noindent\textbf{Hyper-parameter settings for link prediction on KG.}
For each KG, the hyper-parameters are determined by a random search~\cite{bergstra2012random}, including number of layers, learning rate, hidden dimension, batch-size, head number, hop number, teleport probability $\alpha$ and dropout ratios. The hyper-parameter search space is show in Table~\ref{table:randomsearch_linkprediction}.

\begin{table}[!th]
	\centering
	\footnotesize
	\begin{threeparttable}
		\begin{tabularx}{.5\textwidth}{l c c} 
			\hline
			Hyper-parameters & Search Space & Type\\
			\hline
			Initial Entity/Relation Dimension & 100 & Fixed\\
			Number of layers & [$2, 3$] & Choice\\
			Learning rate & [$10^{-4}, 5 \times 10^{-3}$] & Range \\
			Hidden Dimension & [$256, 512, 768$] & Choice \\
			Batch size & [$1024, 2048, 3072$] & Choice \\
			Head Number & [$4, 8$] & Choice \\
			Hop Number & [$2, 3, 4, 5, 6$] & Choice\\
			Teleport probability $\alpha$ & [$0.05, 0.6$] & Range\\ 
			Dropout (attention, feature) & [$0.1, 0.6$] & Range\\
			Weight Decay & [$10^{-10}, 10^{-8}$] & Range\\
			Optimizer & Adam & Fixed\\
			\hline
		\end{tabularx}
	\end{threeparttable}
		\caption{Hyper-parameter search space for link prediction on KG}
	\label{table:randomsearch_linkprediction}
\end{table}

\end{document}